\documentclass[sigconf]{acmart}
\usepackage{graphicx} %
\usepackage{subcaption}
\usepackage{amsmath}
\usepackage[dvipsnames]{xcolor}
\usepackage{listings}
\usepackage{epigraph}

\usepackage{tcolorbox} %
\usepackage{enumitem} %

\usepackage{hyperref}

\renewcommand\footnotetextcopyrightpermission[1]{}
\AtBeginDocument{%
  }

\setcopyright{acmlicensed}
\copyrightyear{2018}
\acmYear{2018}
\acmDOI{XXXXXXX.XXXXXXX}
\acmConference[Conference acronym 'XX]{Make sure to enter the correct
  conference title from your rights confirmation email}{June 03--05,
  2018}{Woodstock, NY}
\acmISBN{978-1-4503-XXXX-X/2018/06}

\begin{document}

\newcommand{\akarsh}[1]{{\color{red}AK: #1}}
\newcommand{\yujin}[1]{{\color{orange}YT: #1}}
\newcommand{\ryan}[1]{{\color{purple}RB: #1}}
\newcommand{\phil}[1]{{\color{blue}PI: #1}}
\newcommand{\ps}[1]{{\color{Mahogany}PS: #1}}
\newcommand{\todo}[1]{{\color{teal}[\textbf{TODO}: #1}]}

\title{Digital Red Queen:\\Adversarial Program Evolution in Core War with LLMs}

\author{
  Akarsh Kumar\textsuperscript{1,2} \quad
  Ryan Bahlous-Boldi\textsuperscript{1} \quad
  Prafull Sharma\textsuperscript{1} \quad \\
  Phillip Isola\textsuperscript{1} \quad
  Sebastian Risi\textsuperscript{2} \quad
  Yujin Tang\textsuperscript{2} \quad
  David Ha\textsuperscript{2} \quad
}

\affiliation{
  \textsuperscript{1}MIT \quad
  \textsuperscript{2}Sakana AI
  \country{}
}

\renewcommand{\shortauthors}{Kumar et al.}

\begin{abstract}
Large language models (LLMs) are increasingly being used to evolve solutions to problems in many domains, in a process inspired by biological evolution.
However, unlike biological evolution, most LLM-evolution frameworks are formulated as static optimization problems, overlooking the open-ended adversarial dynamics that characterize real-world evolutionary processes.
Here, we study Digital Red Queen (DRQ), a simple self-play algorithm that embraces these so-called ``Red Queen'' dynamics via continual adaptation to a changing objective.
DRQ uses an LLM to evolve assembly-like programs, called warriors, which compete against each other for control of a virtual machine in the game of Core War, a Turing-complete environment studied in artificial life and connected to cybersecurity.
In each round of DRQ, the model evolves a new warrior to defeat all previous ones, producing a sequence of adapted warriors.
Over many rounds, we observe that warriors become increasingly general (relative to a set of held-out human warriors).
Interestingly, warriors also become less behaviorally diverse across independent runs, indicating a convergence pressure toward a general-purpose behavioral strategy, much like convergent evolution in nature.
This result highlights a potential value of shifting from static objectives to dynamic Red Queen objectives.
Our work positions Core War as a rich, controllable sandbox for studying adversarial adaptation in artificial systems and for evaluating LLM-based evolution methods.
More broadly, the simplicity and effectiveness of DRQ suggest that similarly minimal self-play approaches could prove useful in other more practical multi-agent adversarial domains, like real-world cybersecurity or combating drug resistance.

\noindent \textbf{Website: } \href{https://pub.sakana.ai/drq}{https://pub.sakana.ai/drq}

\noindent \textbf{Code: }\href{https://github.com/SakanaAI/drq}{https://github.com/SakanaAI/drq}

\end{abstract}


\begin{teaserfigure}
  \begin{subfigure}[t]{0.49\textwidth}
   \captionsetup{labelformat=empty}  %
    \centering
    \includegraphics[width=\textwidth]{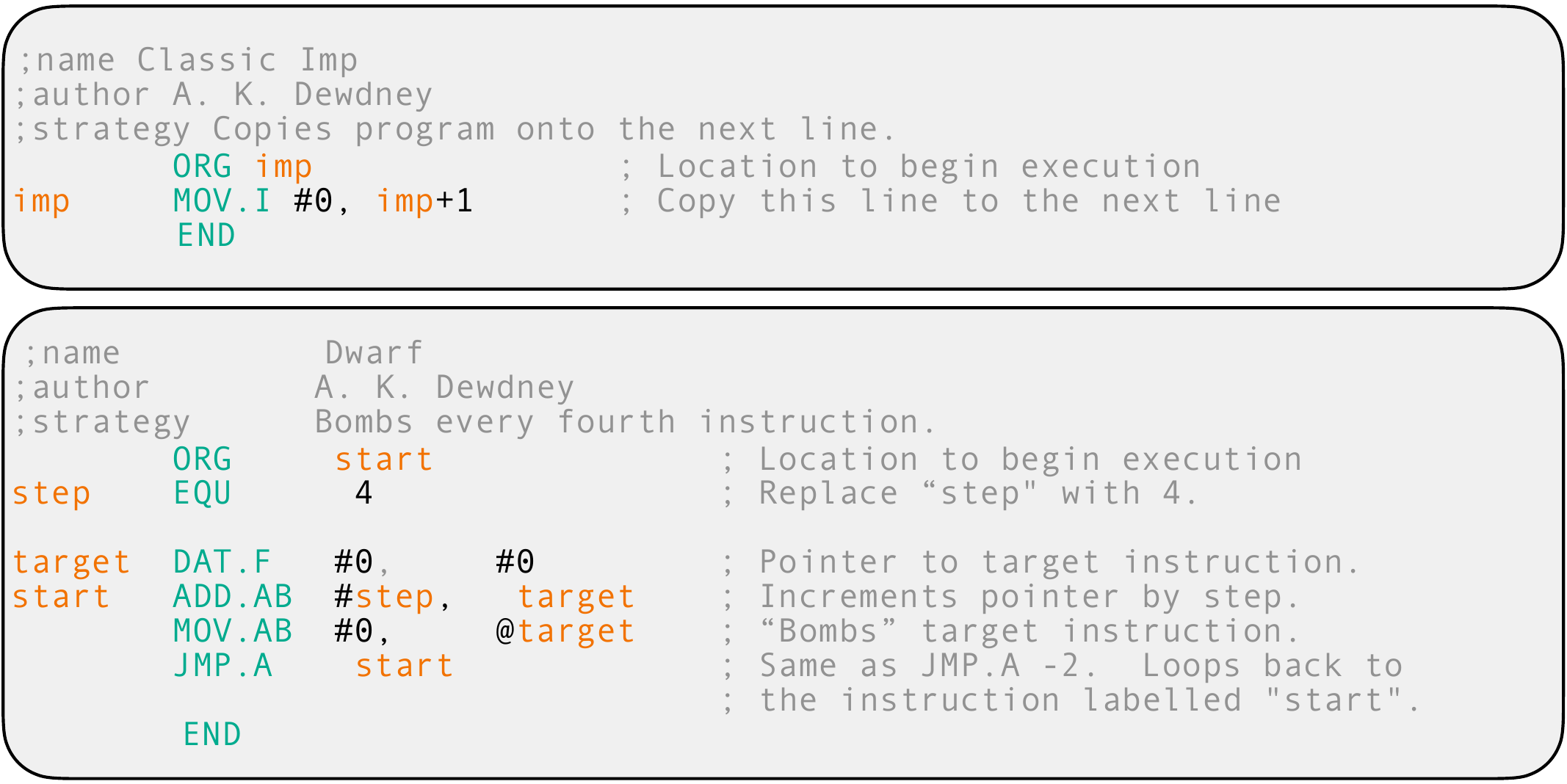}
  \end{subfigure}
  \hfill
  \vspace{7pt}
  \begin{subfigure}[t]{0.49\textwidth}
   \captionsetup{labelformat=empty}  %
    \centering
    \includegraphics[width=\textwidth]{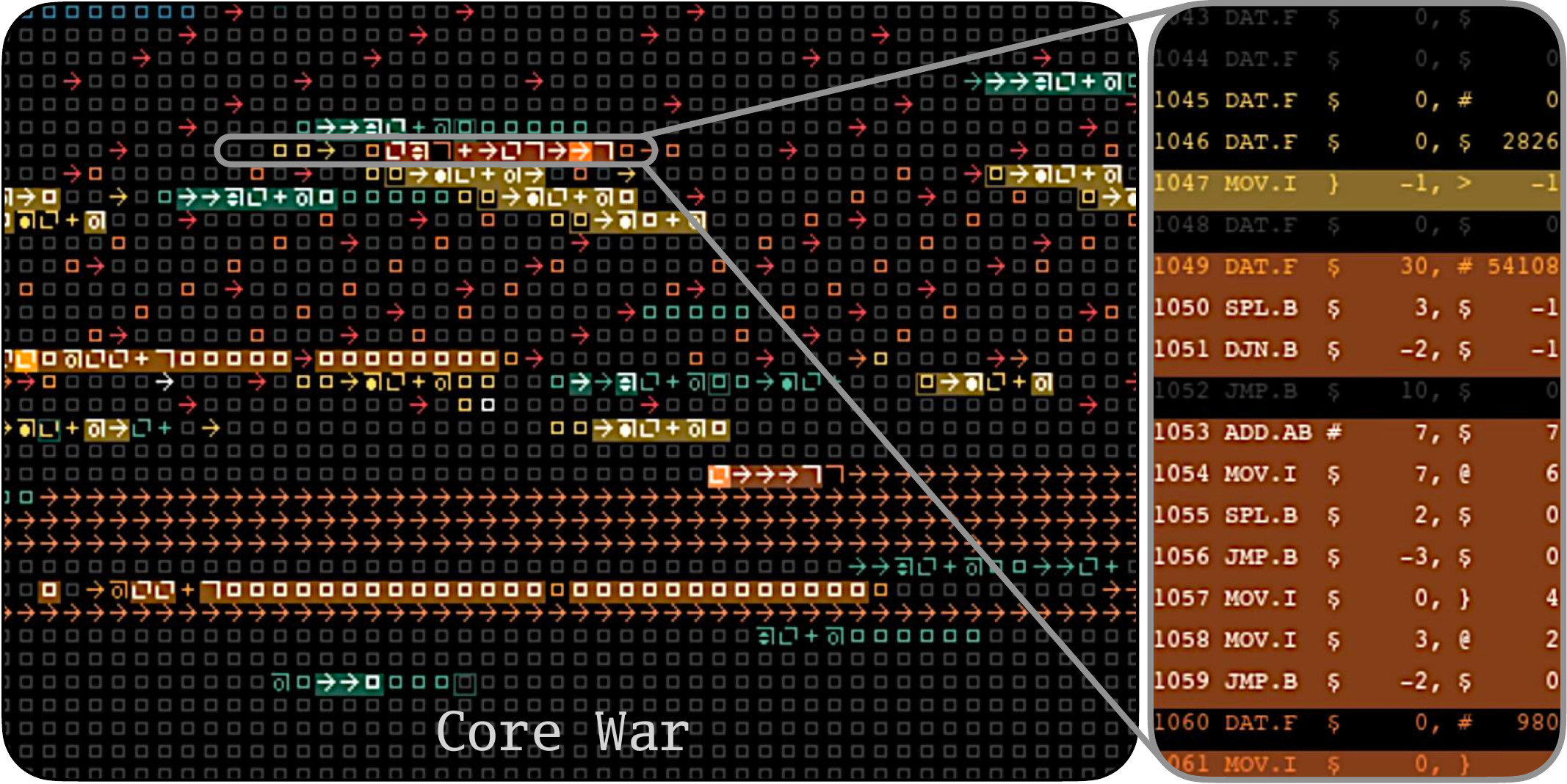}
  \end{subfigure}
  \begin{subfigure}[t]{0.49\textwidth}
   \captionsetup{labelformat=empty}  %
    \centering
    \includegraphics[width=\textwidth]{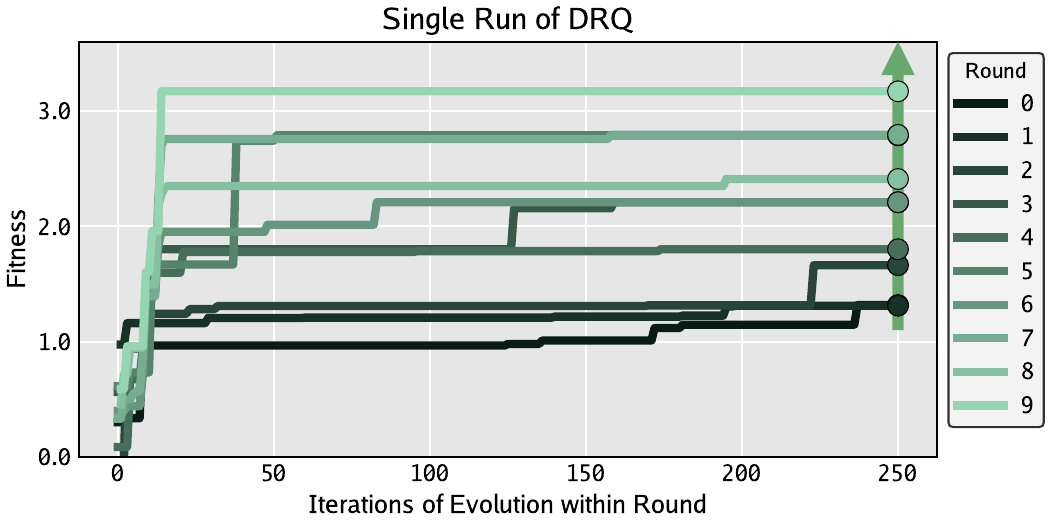}
  \end{subfigure}
  \hfill
  \begin{subfigure}[t]{0.49\textwidth}
   \captionsetup{labelformat=empty}  %
    \centering
    \includegraphics[width=\textwidth]{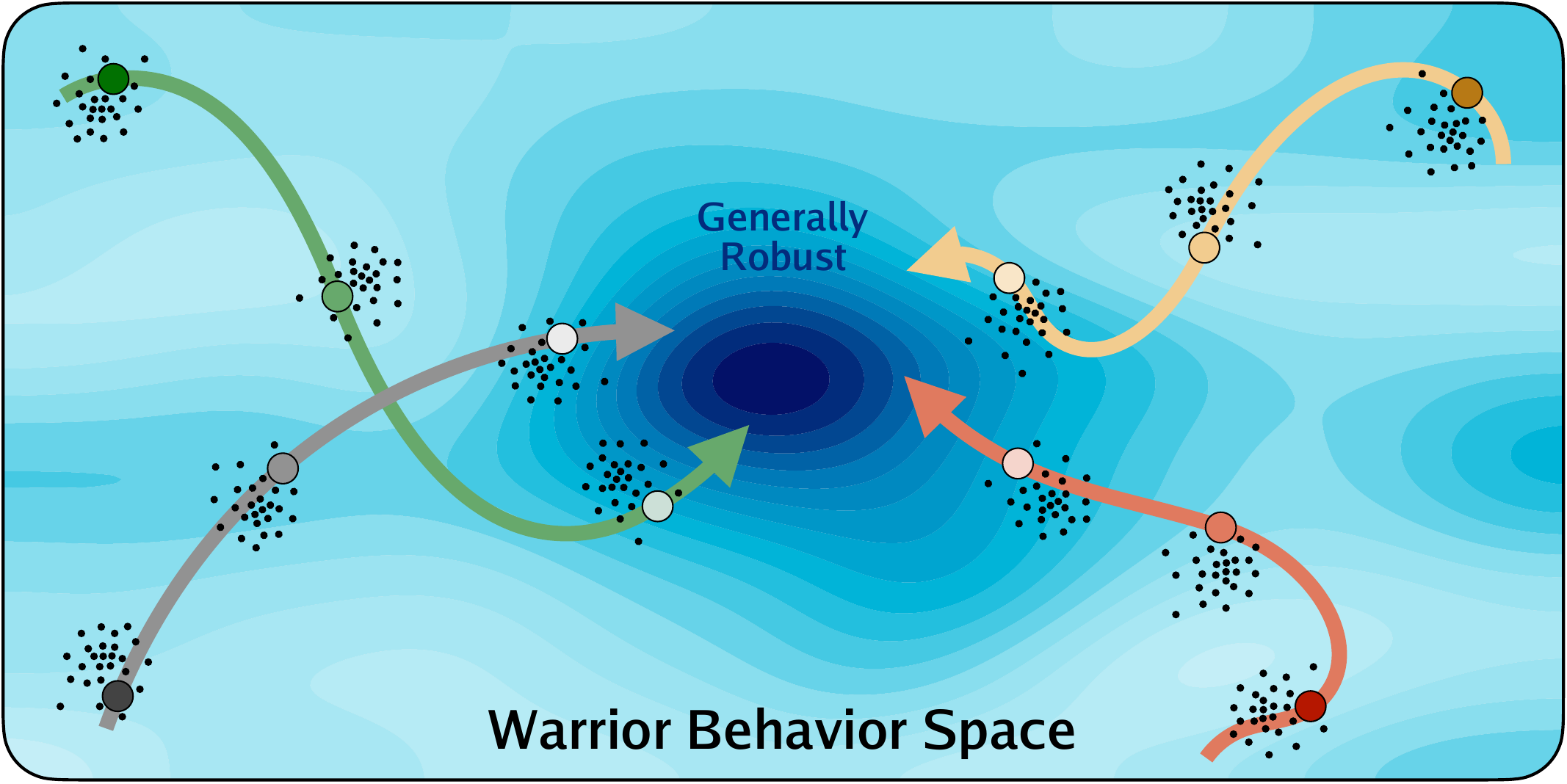}
    \label{fig:teaser-right}
  \end{subfigure}
  \caption{
    \normalfont 
    \textbf{Overview.}
    Digital Red Queen (DRQ), uses large language models (LLMs) to evolve assembly programs called ``warriors'' which compete against each other for control of a virtual machine in the game of Core War.
    \textbf{Top Left:}
    Examples of two classic warriors designed by a human.
    Warriors are written in the Redcode assembly language and then placed into a virtual machine where their instructions are executed alongside other warriors.
    Each warrior seeks to be the last one running by causing opponents to crash.
    \textbf{Top Right:}
    Visualization of the virtual machine’s memory during execution.
    Symbols indicate the instruction opcode, and colors denote the warrior that last modified each address.
    Code and data share the same address space, enabling self-modification and creating a volatile environment for warriors.
    \textbf{Bottom Left:}
    Fitness curves from successive rounds of DRQ.
    Later warriors are trained against all previous warriors.
    \textbf{Bottom Right:}
    A schematic showing that different independent runs of DRQ produce warriors that converge in behavior while becoming generally robust.
  }
  \label{fig:teaser}
\end{teaserfigure}

\maketitle

\setlength\epigraphrule{0pt}
\setlength\epigraphwidth{.3\textwidth}

\section{Introduction}

\epigraph{Now, \emph{here}, you see, it takes all the running you can do, to keep in the same place.}
{Red Queen to Alice\\Lewis Carroll, \emph{Through the Looking-Glass}}

Building on recent test-time scaling trends~\citep{jaech2024openai,snell2024scaling}, researchers are increasingly using large language models (LLMs) to evolve artifacts across many domains~\citep{lehman2023evolution,novikov2025alphaevolve,lange2025shinkaevolve,risi-et-al:book2025}.
By leveraging a grounded selection mechanism~\citep{holland1992adaptation}, these approaches enable LLMs to explore far beyond their pretraining priors~\citep{simon2025lluminate,lehman2023evolution}, making them powerful tools for discovery~\citep{lu2024ai,lange2025towards,novikov2025alphaevolve}.
While recent studies have begun to study open-ended evolution with LLMs~\citep{faldor2024omni,zhang2025darwin,samvelyan2024rainbowteamingopenendedgeneration}, less attention has been devoted to the adversarial dynamics in such evolutionary processes~\citep{dharna2025foundation}.
In this work, we investigate a simple self-play algorithm that uses LLMs to evolve adversarially competing agents.

In the real world, biological, cultural, and technological evolution do not operate as optimization on a static fitness landscape, but are better described as open-ended arms races~\citep{vanvalen1973newevolutionarylaw,dawkins1979arms,stanley2015greatness}.
Solving a single problem is never sufficient: the challenges themselves continually change, whether in the form of foreign viruses developing resistance mechanisms~\citep{irwin2016antiviral} or competitor companies inventing superior products~\citep{schumpeter2013capitalism}.
In evolutionary biology, this continual pressure to adapt is known as the Red Queen hypothesis~\citep{vanvalen1973newevolutionarylaw,cliff_tracking_1995}, which states that organisms must continuously evolve not to gain an \textit{advantage}, but simply to \textit{maintain} their relative fitness in a constantly changing environment.
The name comes from the Red Queen in \textit{Through the Looking-Glass}~\citep{carroll1871through}, whose remark to Alice captures the idea that perpetual adaptation is necessary just to avoid falling behind.
As more LLM systems are deployed into the real world and interact with each other, it is likely that they too will begin to exhibit similar evolutionary Red Queen dynamics~\citep{zhang2025llms}.

To prepare for such a future, it is important to study the Red Queen dynamics of LLMs in an isolated scientific setting.
This pursuit requires a test bed that is rich enough to yield insights relevant to the real world, while still being in a sandbox where the researcher maintains full control.
Simulations from artificial life and cybersecurity naturally lend themselves to this goal since they prioritize adversarial dynamics in controlled environments~\citep{langton1997artificial}.

For this reason, we use \textit{Core War}~\citep{dewdney1984corewar} as a testbed for studying Red Queen dynamics with LLMs.
Core War is a classic programming game, studied in both artificial life~\citep{rasmussen1990coreworld} and cybersecurity~\citep{maliukov2024evolving}, where low-level assembly-like programs, called warriors, compete for control of a shared virtual computer.
To run a battle, the warriors’ raw assembly code is placed into random locations in memory, and the virtual machine executes their instructions line by line.
Because code and data share the same address space, self-modifying logic is common, creating a very volatile environment.
Each warrior aims to be the last one running by causing its opponents to crash while preserving its own survival.
Core War is Turing-complete, making it rich enough to run any computation and, in principle, support an open-ended arms race.
It is also fully bounded within a sandboxed simulator, far removed from real-world consequences.

To study adversarial evolution in Core War, we develop an algorithm called \textit{Digital Red Queen} (DRQ) that uses LLMs to perform multiple rounds of evolutionary optimization to create new warriors.
DRQ is initialized with a single warrior program.
In the first round, it uses an LLM to evolve a second warrior that defeats the initial warrior within the Core War simulation.
In each subsequent round, DRQ continually evolves a new warrior to defeat all previous ones in a multi-agent simulation.
The champions of each round form a sequence of adapted warriors.
Rather than treating DRQ as a fundamentally novel algorithm, it should be viewed as a deliberately minimal instantiation of prior self-play approaches~\citep{heinrich2015fictitious,lanctot2017unifiedgametheoreticapproachmultiagent,dharna2025foundation}, adapted to Core War for scientific study.
Figure~\ref{fig:teaser} provides an overview of the Core War domain, the DRQ method, and the resulting evolutionary dynamics.

DRQ uses MAP-Elites~\citep{mouret2015illuminating}, a quality-diversity algorithm, to optimize warriors within each round, preventing diversity collapse during search.
By playing against all previous round champions, DRQ avoids cyclic adaptations across rounds, consistent with techniques in prior work~\citep{zhang2024survey}.
Together, these design choices allow DRQ to unlock the power of self-play for generating entities that compete in simulated environments.

We evaluate DRQ by putting the generated warriors in competition against human-designed warriors.
The baseline of static optimization (single-round DRQ) is able to synthesize specialist warriors that \textit{collectively} \textbf{defeat or match} $283$ out of $294$ human warriors.
However, inspection of individual performance reveals that these warriors are brittle and overfit to their training opponent: any single warrior defeats only about $28\%$ of the human-designed warriors.
In contrast, DRQ trains against a growing history of opponents, implicitly incentivizing the emergence of robust, generalist strategies capable of handling diverse threats.
When running full DRQ, analysis of its sequence of warriors reveals an intriguing pattern: with more DRQ rounds, the resulting warriors become \textbf{increasingly general}, while simultaneously exhibiting \textbf{reduced behavioral diversity} across independent runs.
Together, these two trends indicate an emergent convergence pressure toward a single general-purpose behavior in Core War.
This phenomenon is reminiscent of convergent evolution in nature, such as the independent evolution of mammalian and insect eyes to address similar functional demands.

Given the strong performance of DRQ in Core War, we investigate the extent to which LLMs understand this domain.
The mapping from a warrior’s Redcode source code to its performance requires an expensive simulation that is highly chaotic: small changes in code can lead to drastic changes in battle outcomes.
Given the large number of warriors generated by DRQ, we ask whether an LLM can directly predict the outcome of a battle between two warriors using only their source code.
To test this, we embed warrior source code using a text embedding model and train a linear probe to predict the warrior's final generality score.
We are able to predict generality scores with a test $R^2=0.461$.
This result opens a path toward strengthening such predictors and ultimately using them either as surrogate models to bypass simulation or as interpretability tools for understanding what makes source code effective.

Overall, DRQ illustrates how LLM research might move beyond static problem settings and toward more realistic open-ended environments characterized by Red Queen dynamics.
At the same time, we hope this work encourages adoption of Core War as (1) a safe and expressive testbed for studying artificial evolution and (2) a benchmark for evaluating an LLM's Red Queen capabilities.
The DRQ algorithm itself is simple and general, and could be applied to other adversarial domains, such as discovering real cybersecurity exploits/defenses, designing biological viruses/drugs, or exploring any other complex multi-agent environment of interest.
Systematically exploring adversarial dynamics in controlled environments is an important step toward discovering potential dangers before they arise in real-world systems.

\section{Related Work}

\paragraph{Program-Based Competition in Artificial Life}
Driven by the goal of understanding ``life as it could be''~\citep{langton1992artificial}, artificial life (ALife) research has historically explored ecosystems of programs that compete with one another.
Tierra~\citep{ray1991evolution} instantiated one of the first such environments: it featured a shared memory ``soup'' where self-replicating machine code competed for CPU cycles, leading to the emergence of complex ecological phenomena like parasitism.
Avida~\citep{ofria2004avida} extended this line of work by introducing a structured 2D lattice and private address spaces for each organism, allowing for the evolution of complex logic tasks through a system of computational rewards.
A more recent study showed how spontaneous self-replication can emerge with a very minimal language~\citep{alakuijala2024computational}.
These works show that program environments can produce life-like phenomena, making them compelling test beds to study evolution.

\paragraph{Core War}

Core War was originally made as a competitive programming game in 1984~\citep{dewdney1984corewar,jones1984corewar} and continues to fascinate researchers and hobbyists as a microcosm of digital evolution and adversarial computation.
In this game, players write programs called warriors in Redcode, an assembly-like language designed for the Core War simulator.
Warriors are loaded into a virtual computer’s memory array, known as the Core, where they battle for control of the system.
The Core is a circular memory of fixed size (typically 8{,}000 cells) each containing one instruction. 
Each warrior is granted one process at initialization, represented by a program counter indicating the next address to execute.
Each simulation step, the virtual machine executes one address per warrior in a round-robin fashion.

Because the Core does not distinguish between code and data, every instruction can be read, written, or executed. 
This creates a highly volatile environment where self-modifying code is commonplace.
A program can inject a \texttt{DAT} instruction in front of an opponent's process, terminating it when that process attempts to execute it.
Some strategies include \textit{bombing} (placing \texttt{DAT}s throughout the Core), \textit{replication} (copying the warrior’s own code into multiple memory locations), and \textit{scanning} (probing the Core to locate enemies before striking)~\citep{corewar_strategies_docs}. 
These strategies interact dynamically, creating an ecosystem of possibilities.

Many prior works have evolved warriors using genetic programming~\citep{corno2004evolution,tasnim2016evolving,andersen2001garden,corno2003exploiting,sanchez2006evolving,perry1991cwgenetics,corno2005evolving}.
For instance, \citet{corno2005evolving} used their $\mu$GP framework to evolve programs, producing some of the top-performing ``nano'' warriors.
However, most of these approaches were only effective on small Core sizes and did not scale well to the full Core War environment.
None of these methods leverage LLMs or investigate the evolution dynamics at a large scale.

\paragraph{Open-Ended Coevolution and Self-Play}

One of the core mechanisms that led to complexity in biology is the Red Queen dynamics of evolutionary arms races~\citep{ridley1994red}.
Accordingly, researchers have tried to capture this mechanism in silico through coevolution between populations of agents~\citep{paredis1995coevolutionary}.
Polyworld~\citep{yaeger1994computational} evolved neural-network–controlled agents in a simulated environment, where agents competed for resources and reproduced, developing behaviors such as predation and cooperation.
POET \citep{wang2019pairedopenendedtrailblazerpoet,wang2020enhancedpoetopenendedreinforcement} co-evolves agents and their environments, open-endedly generating problems and their solutions.
Many other works have featured coevolution as a primary mechanism to get complexity~\citep{parker2022evolving,lu2024jaxlife,dharna2020co,stanley2004competitive,brant2017minimal}.
Quality-diversity algorithms have been proposed as a way to stabilize evolutionary algorithms by maintaining diversity~\citep{lehman2011abandoning,lehman2011evolving,mouret2015illuminating,anne2025generational}.

Reinforcement learning has also taken inspiration from Red Queen dynamics in the form of self-play.
Self-play describes situations in which agents are trained in environments where the opponent is themselves, a historical copy of themselves, or related to them in some way~\citep{zhang2024survey}.
Self-play has driven some of the most significant breakthroughs in AI, ranging from early successes in checkers~\citep{samuel1959some} and backgammon~\citep{tesauro1995temporal} to modern advances in board games such as Go~\citep{silver2016mastering,silver2017mastering}, chess~\citep{silver2017mastering_zero}, and 3D multi-agent environments~\citep{bansal2017emergent, baker2019emergent}.
It has also been a key factor in mastering complex real-time multiplayer strategy games like StarCraft II~\citep{vinyals2019grandmaster,arulkumaran2019alphastar} and Dota 2~\citep{berner2019dota}.
Recent work has also shown that self-play can create robust self-driving car policies~\citep{cusumano2025robust}.
Self-play can be viewed as a form of automatic curriculum learning~\citep{portelas2020automatic}, and can thus be abstracted as one agent interacting with the environment while another generates the environment~\citep{dennis2020emergent}.

Within self-play, our DRQ algorithm is closely related to Fictitious Self-Play (FSP)~\citep{brown1951iterative,heinrich2015fictitious,heinrich2016deep} and Policy Space Response Oracles (PSRO)~\citep{lanctot2017unifiedgametheoreticapproachmultiagent}, which provide game-theoretic frameworks for multi-agent learning.
FSP trains agents by learning approximate best responses to the empirical average of their opponents’ past policies.
PSRO iteratively expands a population of policies by training approximate best responses to mixtures of existing strategies and solving a meta-game to compute Nash equilibrium distributions over the strategy population.
In contrast, DRQ does not construct explicit meta-strategies or solve a meta-game; instead, we directly optimize the current agent within a multi-agent environment containing all previous agents.
Furthermore, because our domain lacks a well-defined action space, we employ an evolutionary algorithm in the inner loop to optimize agents, allowing our approach to extend beyond standard action-based game settings.
Finally, we use LLMs to guide the evolutionary optimization process.

\paragraph{LLM-Guided Evolution}

Recent work has begun to merge LLMs with evolutionary algorithms, using LLMs as intelligent mutation or generation operators.
This approach is appealing because it exploits the model’s prior knowledge to propose domain-aware edits, while grounded selection expands discovery capabilities beyond the model’s pretraining distribution~\citep{lehman2023evolution}.
\citet{lehman2023evolution} was the first to show that evolution with LLM-driven mutations can create solutions for out-of-distribution tasks.
AlphaEvolve~\citep{novikov2025alphaevolve} showed that scaling this approach results in a general system capable of discovering new breakthroughs in a variety of domains, including logistical scheduling, hardware chip design, and efficient matrix multiplication algorithms.
Many other works have used LLM-guided evolution for everything from prompt optimization to self-referential agentic code~\citep{fernando2023promptbreederselfreferentialselfimprovementprompt,romera2024mathematical,lange2025shinkaevolve,zhang2025darwin,zhang2023omni,faldor2024omni,ma2023eureka,lu2024ai,lange2025towards,liang2024eurekaverse}.
These works demonstrate that LLM-guided evolution can act as engines of discovery in many domains.

\paragraph{LLMs for Self-Play}
\citet{dharna2025foundation} was the first to propose Foundation Model Self-Play (FMSP), which uses LLMs to create agent policies that compete against each other in two-player games such as a 2D evader–pursuer game and an LLM red-teaming game.
DRQ differs from FMSP both in its technical algorithmic details (each new agent is optimized in an environment containing all previous agents) and in its application domain (Core War is a richer domain that directly simulates a computer).
Additionally, our work focuses more on the science of evolutionary self-play rather than proposing a specific method.

\citet{Bachrach2025LLMPSRO} uses LLMs within PSRO to generate Checkers agents.
Our work differs heavily in algorithmic details, application domain, and motivation.

Self-play with LLMs has also been used to improve LLM capabilities~\cite{wei2025toward,cheng2024self,chen2025spc,liu2025spiral,wang2025battling,yuan2025marshal,liu2025spice}.

Our work unifies these threads by connecting LLMs, coevolution, self-play, and ALife within the rich testbed of Core War.
This combination enables the study of Red Queen dynamics in a controlled, yet expressive, environment.

\section{Methods: Digital Red Queen}
\label{sec:methods}

Our approach, which we call \textit{Digital Red Queen} (DRQ), is built on prior works on self-play~\citep{heinrich2015fictitious,lanctot2017unifiedgametheoreticapproachmultiagent,dharna2025foundation} and coevolutionary training~\citep{hillis_co-evolving_1990,rosin_new_1997,tang2020learning}, adapted for the Core War setting.

\paragraph{DRQ Algorithm}
DRQ begins with an initial warrior $w_0$ and proceeds through a sequence of $T$ rounds, each performing an evolutionary optimization.
In each round $t$, a new warrior $w_t$ is evolved to defeat the set of all previous warriors $\{ w_0, w_1, \ldots, w_{t-1} \}$.
This process induces a competitive pressure that changes every round, driving the emergence of novel strategies and counter-strategies.
The algorithm is detailed below: 
\begin{tcolorbox}
\begin{enumerate}[leftmargin=1.5em]
    \item \textbf{Initialization:} Start with a base warrior $w_0$ which is either human designed or LLM-generated.
    \item \textbf{Adversarial Optimization:} At round $t$, optimize a new warrior $w_t$ to maximize its expected fitness in an environment which includes all prior warriors:
    \[
    w_t = \arg\max_w \; \mathbb{E} \left[ \text{Fitness}(w; \quad \{w_0, \dots, w_{t-1})\}\right]
    \]
    The expectation is over different seeds of evaluation.
    
    \item \textbf{Iteration:} Repeat for $T$ rounds, generating a lineage of warriors $\{w_0, w_1, \ldots, w_T\}$.
\end{enumerate}
\end{tcolorbox}

We do not update older warriors in the lineage, as prior work has shown that historical self-play promotes stability and mitigates cyclic dynamics~\citep{zhang2024survey}.

Because the number of warriors increases each round, the marginal influence of any newly introduced warrior on the environment decreases over time, implying that the induced fitness function changes less and less as $T\to\infty$.

Since program synthesis presents a highly deceptive search landscape, most greedy algorithms can get stuck in local minima~\citep{koza1994genetic,real2020automl}.
This motivates the use of MAP-Elites~\citep{mouret2015illuminating}, a diversity preserving algorithm, as our choice of optimization within a round.
We later provide empirical evidence that diversity preservation is indeed beneficial in Core War.

\paragraph{Intra-round Optimization with MAP-Elites}
MAP-Elites is a widely used quality-diversity algorithm that discretizes a user-defined behavioral descriptor space into a set of cells, each storing at most one elite solution that exhibits the behavioral characteristics of that cell.
By restricting competition to solutions that fall within the same cell, MAP-Elites imposes localized selection pressure while preserving global diversity.
Partitioning with respect to behavior allows the archive to maintain a broad set of stepping stones, many of which may be individually poor but crucial for discovering strong strategies in other regions of behavior space.
This property makes MAP-Elites particularly well suited for Redcode program synthesis.%

MAP-Elites follows a simple evolutionary procedure.
The archive $\mathcal{A}$ maps a predefined set of behavioral cells $\mathcal{C}$ to their current elite solutions.
After initializing $\mathcal{A}$ with random solutions, it performs the following steps:
(i) randomly sample an individual $w$ from $\mathcal{A}$;
(ii) mutate $w$ to produce an offspring $w'$;
(iii) evaluate its fitness $f=\text{Fitness}(w'; \dots)$ and behavioral descriptor cell $c=\text{BD}(w') \in \mathcal{C}$;
(iv) insert $w'$ into the archive at $c$ if $f$ exceeds the fitness of the current elite in $c$ (or if that cell is empty).
Iterating this process gradually fills the archive with increasingly high-performing behaviorally diverse solutions.

The fitness function depends on the current round of DRQ $t$, yielding $\text{Fitness}(\cdot, \{w_0, \dots, w_{t-1}\})$.
We define the behavior descriptor function $BD(\cdot)$ as the discretized tuple (total spawned processes, total memory coverage), which captures two high-level aspects of a warrior’s behavior during simulation.
We optionally initialize the MAP-Elites archive in round $t$ using all previous champions $\{w_0, \dots, w_{t-1}\}$ to bootstrap the optimization.

\paragraph{LLMs as the Mutation Operator}

Within DRQ, LLMs are used to generate new warriors and to mutate existing ones.
In all cases, the model receives a system prompt describing the Core War environment and a concise manual for the Redcode assembly language, including its opcodes, addressing modes, and an example warrior.
To generate a new warrior, the LLM is given a user prompt instructing it to produce a novel Redcode program.
To mutate an existing warrior, the LLM is provided with the original program and instructed to modify it in ways that could improve performance.
See Appendix~\ref{sec:appendix_prompts} for details on the prompts used in DRQ. 

We intentionally chose this simplistic use of LLMs to keep the focus of the study on Core War and the analysis of evolution, rather than on LLM-specific techniques.
Other methods for applying an LLM to modify code exist and could easily be integrated into the DRQ framework.
For example, an LLM could output a diff~\citep{piterbarg2023diff}, or it could be conditioned on the results of the simulation to provide more informative feedback~\citep{shinn2023reflexion}.

It is possible to run DRQ without LLMs by relying solely on random generation and random mutation over the space of opcodes, addressing modes, and numeric parameters.
However, in extremely sparse search spaces, where most points and mutations produce invalid or non-functional programs, some prior over the search space is crucial for practical search efficiency~\citep{real2020automl}.
LLM-based priors can speed up search by orders of magnitude~\citep{austin2021program}.

\paragraph{Self-Play and Red Queen Dynamics}
DRQ is purposely one of the simplest multi-agent self-play algorithms that can be constructed for evolving warriors in Core War.
DRQ's multi-round design ensures that fitness is not measured by performance against a fixed opponent, but rather against a continually growing population of opponents.
This shifting landscape embodies Red Queen dynamics: each new warrior must continually adapt to overcome the latest strategies, driving a process of adversarial innovation.

\section{Experiments}

We evaluate DRQ with experiments designed to assess 1) its ability to evolve generally competitive Core War programs, and 2) its capacity for continual improvement through Red Queen dynamics.

All experiments use the following fitness function, which accounts for both survival and dominance within the battle.
In a battle with $N$ warriors and $\mathcal{T}$ simulation timesteps, a total of $N$ units of fitness are distributed evenly over time.
At each timestep, the remaining (living) warriors share a fitness of $N/\mathcal{T}$.
This design incentivizes warriors to survive as long as possible while also eliminating others to increase their share of the reward.
The cumulative reward across all timesteps defines the warrior’s fitness:
$$ \text{Fitness}(w_i; \quad \{w_j\}_{j\neq i}) = \sum_{\tau=1}^{\mathcal{T}} \frac{N}{\mathcal{T}} \frac{A^i_\tau}{\sum_j A^j_\tau} $$
where $A^i_\tau$ is an indicator for whether warrior $i$ is alive at simulation timestep $\tau$.
Note that a warrior's fitness is context-dependent on other warriors.
A warrior is said to defeat another warrior if it achieves higher fitness in a 1-on-1 battle between the two.

All experiments use the following MAP-Elites behavioral descriptors: 1) the total number of spawned threads (via \texttt{SPL} opcodes), and 2) the total memory coverage of the warrior during simulation.
These two axes capture two important strategical aspects of warriors in Core War.
The grid is discretized in log space.

All experiments use GPT-4.1 mini (\texttt{gpt-4.1-mini-2025-04-14}) \citep{openai_api_2025} as the LLM.
Preliminary experiments did not show significant performance increase with larger models.

For terminology, \textit{rounds} correspond to steps of DRQ (outer loop), while \textit{iterations} correspond to optimization steps within a round (inner loop).
More experimental details are provided in Appendix~\ref{sec:appendix_corewar_details}.

\subsection{Static Target Optimization Against Human Warriors}

The first experiment evaluates the effectiveness of static optimization against a target.
This baseline corresponds to a single round of DRQ.
We use a dataset of $294$ human warriors and perform one $1000$-iteration optimization run for each.
We do not initialize the optimization with the human warriors.

\begin{figure}
    \centering
    \vspace{-0.3cm}
    \includegraphics[width=1.0\columnwidth]{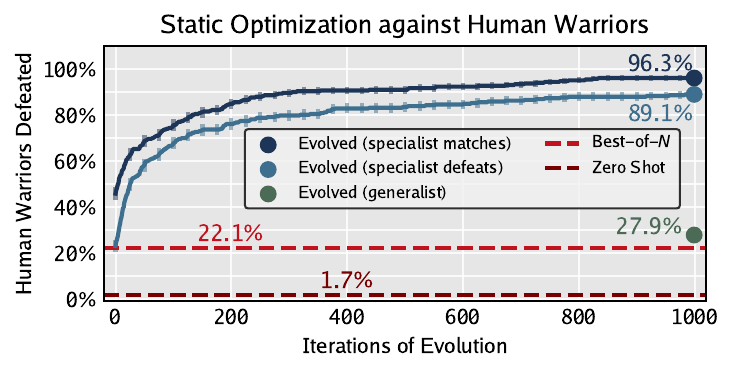}
    \vspace{-0.7cm}
    \caption{
    \normalfont 
    \textbf{Static optimization baseline.}
    Static optimization (single-round DRQ) with an LLM can discover specialist warriors that collectively match or surpass $96.3\%$ of $294$ human-designed warriors, far above the LLM's zero-shotting and best-of-$N$ baselines.
    However, individual warriors are brittle, defeating or matching only $27.9\%$ of human-designed warriors on average.
    }
    \label{fig:main}
\end{figure}

Figure~\ref{fig:main} summarizes the results.
A single warrior generated by the LLM zero-shot defeats, on average, only $1.7\%$ of all human warriors, which is expected given that Redcode is relatively out-of-distribution in most LLM pretraining datasets.
Using a best-of-$N$ sampling strategy produces a set of warriors that can collectively defeat $22.1\%$ of human warriors for $N=8$.
In contrast, evolutionary optimization against each human warrior generates a specialized warrior for every opponent; this set can collectively defeat $89.1\%$ of human warriors and defeat or tie $96.3\%$.
The large jump in performance from best-of-$N$ to evolved warriors demonstrates how evolution can drive performance in out-of-distribution domains.

These numbers reflect \textbf{specialist} performance: the percentage of human warriors defeated by \textit{at least one} of the evolved warriors.
Another metric is \textbf{generalist} performance: the percentage of human warriors defeated or tied by a \textit{single} warrior.
On average, an evolved warrior can defeat or tie only $27.9\%$ of all human warriors, indicating that they are brittle and likely overfit to their training opponent.

\subsection{Iterative Red Queen Dynamics}

\begin{figure*}[!h]
    \centering
    \includegraphics[width=1.0\textwidth]{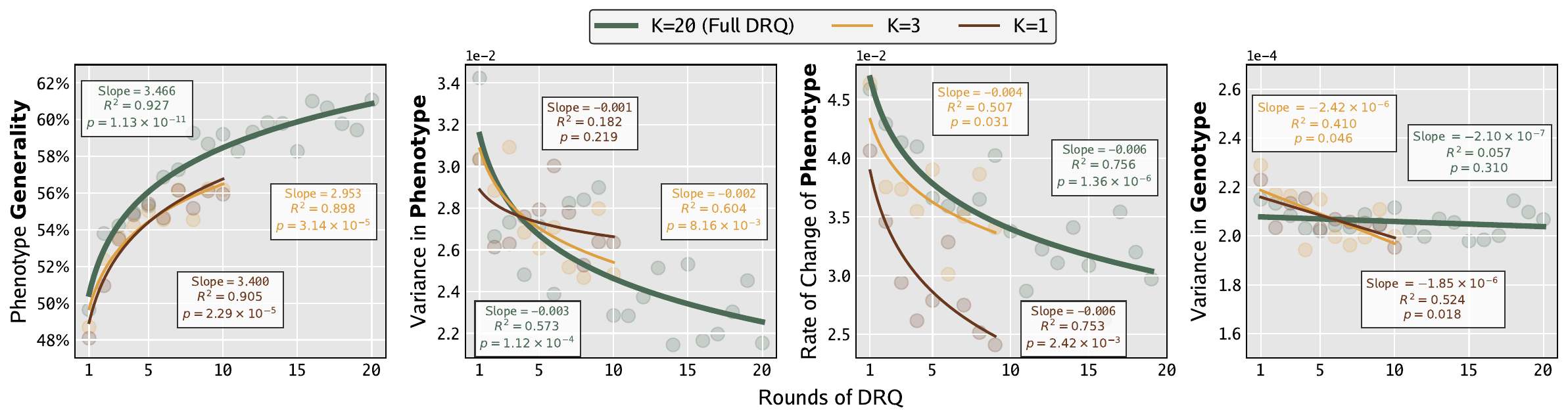}
    \vspace{-0.7cm}
    \caption{
    \normalfont 
    \textbf{DRQ warriors are statistically converging toward a single general-purpose behavior over rounds.} 
    Each point in all plots is computed from 96 independent DRQ runs with different initial warriors.
    Logarithmic or linear models are fit to the data, and reported $p$-values test the null hypothesis that the slope of the fitted model is zero.
    $K$ is the history length in DRQ.
    \textbf{Left:}
    The warriors' average generality increases over rounds.
    Generality is defined as the fraction of unseen human warriors defeated or tied, measuring a warrior’s ability to adapt to novel threats in a zero-shot setting.
    \textbf{Center Left:}
    The variance of the warriors’ phenotype across independent DRQ runs decreases over rounds.
    A warrior’s phenotype is defined as a vector of fitness values against each unseen human warrior.
    \textbf{Center Right:}
    The rate of change of the phenotype decreases over rounds.
    Under the log model, full convergence would require an exponential number of rounds.
    \textbf{Right:}
    The variance of the warriors’ genotype across independent DRQ runs remains static over rounds.
    A warrior’s genotype is defined as a text embedding of its source code.
    }
    \label{fig:generality_vs_rounds}
    \vspace{-0.3cm}
\end{figure*}

Our second experiment investigates the dynamics of running DRQ for multiple rounds.
Due to the computational cost, we select a smaller dataset of $96$ diverse human warriors and conduct multi-round DRQ runs against each one.
We ablate the effect of history length $K$ in DRQ, which determines how many previous warriors each round optimizes against.
For example, $K=1$ plays against only the previous round’s champion, while $K=3$ considers the champions from the previous three rounds.
We initialize the optimization in each round with all prior champions.

To ground the analysis, for each query warrior we measure its fitness in 1-on-1 battles against a dataset of $317$ human warriors.
A warrior’s \textbf{generality} is defined as the fraction of human warriors it defeats or ties, measuring its robustness to new threats in a zero-shot manner.
A warrior’s \textbf{phenotype} is defined as the vector of fitness values against each unseen human opponent, capturing its black-box performance profile against a diverse range of strategies.
A warrior’s \textbf{genotype} is defined as a text embedding of its source code, representing the lowest-level description of the warrior.
We get embeddings using the OpenAI \texttt{text-embedding-3-small} model~\citep{openai_api_2025}.
Similar to real biology, different genotypes may correspond to similar phenotypes, and small changes in genotype can induce large changes in phenotype.

Figure~\ref{fig:generality_vs_rounds} summarizes the dynamics of multi-round DRQ across 96 independent runs.
Across all history lengths $K$, we observe a consistent increase in average generality over rounds (Figure~\ref{fig:generality_vs_rounds}, Left), indicating that DRQ progressively discovers more robust warriors.
This trend suggests that optimizing against a small but changing set of adversaries can induce a pressure towards generality.

At the phenotype level, DRQ exhibits two distinct forms of convergence.
First, the variance of warriors’ phenotypes across independent runs decreases over rounds (Figure~\ref{fig:generality_vs_rounds}, Middle Left), indicating convergence \textit{across different initial conditions}.
Second, the rate of change of the phenotype decreases over rounds within each run (Figure~\ref{fig:generality_vs_rounds}, Middle Right), indicating convergence toward a stable phenotype \textit{within a single run}.
The latter effect is partly expected, as the fitness function defined in Section~\ref{sec:methods} changes more slowly in later rounds.
However, convergence across different independent runs is largely unexpected and suggests a universal attractor in phenotype space.

\begin{figure}[t]
    \centering
    \includegraphics[width=1.0\columnwidth]{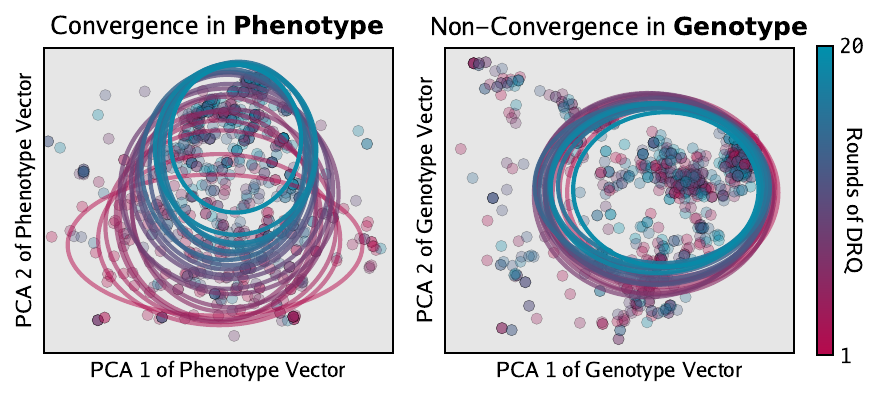}
    \vspace{-0.5cm}
    \caption{
    \normalfont 
    \textbf{Convergence is observed in the phenotype but not in the genotype.}
    Moreover, this convergence pressure is relatively weak and does not appear in every DRQ run, but only emerges statistically when aggregating many independent runs.
    }
    \label{fig:convergence_pca}
\end{figure}

In contrast, no corresponding convergence is observed at the genotype level.
The variance of genotypes across runs remains approximately constant over many rounds (Figure~\ref{fig:generality_vs_rounds}, Right), indicating that DRQ does not collapse onto a single canonical implementation.
This dissociation between phenotypic and genotypic convergence is further emphasized in Figure~\ref{fig:convergence_pca}, which visualizes two principal axes of the phenotype and genotype spaces.

Under the logarithmic fits in Figure~\ref{fig:generality_vs_rounds}, full phenotypic convergence would require an exponential number of rounds, implying that while convergence pressure exists, it is weak and only detectable statistically when aggregating many runs (Figure~\ref{fig:convergence_pca}).

\begin{figure}[b]
    \centering
    \includegraphics[width=1.0\columnwidth]{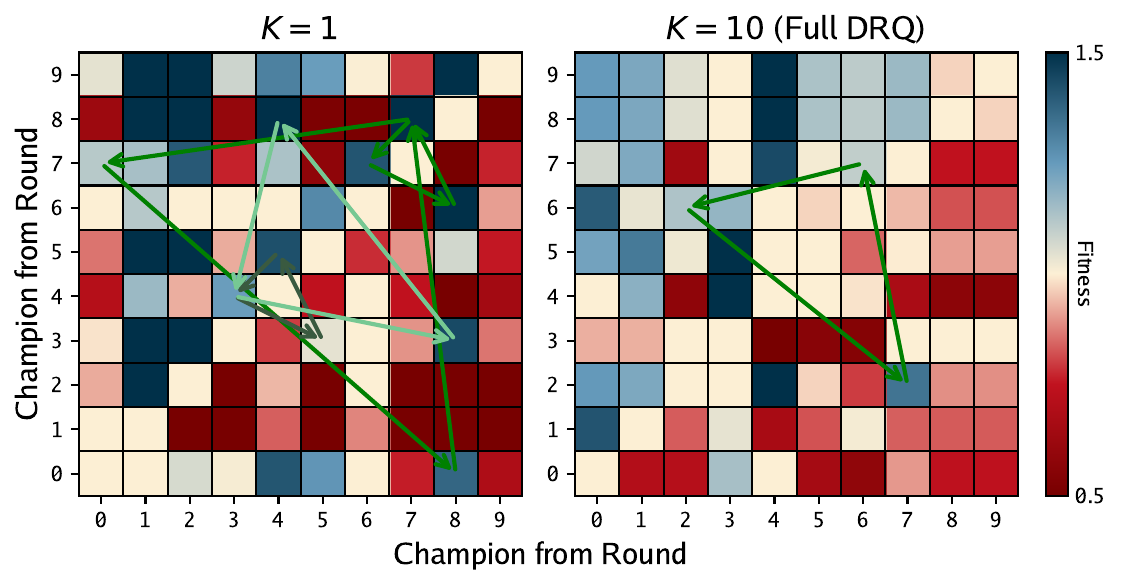}
    \caption{
    \normalfont 
    \textbf{Cyclic behavior in DRQ across champions from different rounds.}
    Arrows show cycles of three warriors that have a rock-paper-scissors dynamic.
    DRQ with $K=1$ exhibits many cycles, whereas full DRQ reduces them.
    }
    \label{fig:cycles}
\end{figure}

Taken together, these results suggest that DRQ drives warriors toward similar general-purpose behaviors while preserving diversity in their underlying implementations.
This mirrors the phenomenon of convergent evolution in biology: different species have evolved similar traits (like eyes or wings) independently, but through distinct genetic mechanisms.
In both DRQ and biology, this phenomenon is likely driven by the selection pressure on phenotypic function rather than on the underlying genotypic representation.

\begin{figure}[b]
    \centering
    \includegraphics[width=0.7\columnwidth]{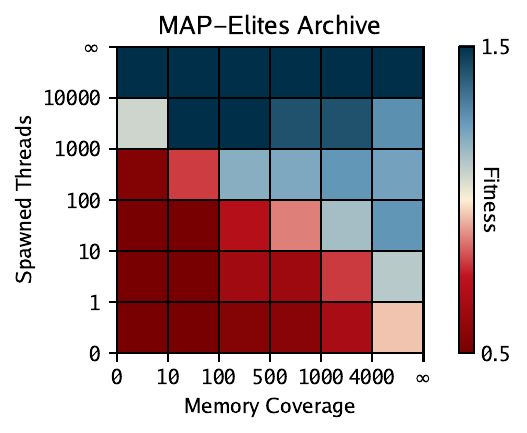}
    \vspace{-0.3cm}
    \caption{
    \normalfont 
    \textbf{MAP-Elites archive with fitness averaged across rounds and runs.}
    The axes correspond to memory coverage (total number of unique addresses accessed during execution) and spawned threads (total number of threads launched via a fork opcode)
    The best warriors have low memory coverage and many spawned threads.
    }
    \label{fig:me-archive}
\end{figure}

\subsection{Cyclic Dynamics}

Cyclic dynamics are a well-known phenomenon in self-play and coevolutionary systems, where agents rotate among strategies that dominate one another, analogous to rock–paper–scissors~\citep{watson_coevolutionary_2001,ficici1998challenges}.
Such dynamics have also been observed in Core War~\citep{corewar_strategies_docs}.
In this section, we analyze the extent to which DRQ has cyclic behaviors.

We define a cycle as a triplet of warriors $(a, b, c)$ such that $a$ defeats $b$, $b$ defeats $c$, and $c$ defeats $a$.
As the history length increases from $K=1$ to $K=10$ (full DRQ), we observe a $77\%$ reduction in the total number of cycles across all runs.
This finding is consistent with prior work showing that incorporating historical opponents into self-play reduces cyclic behavior~\citep{rosin_new_1997,vinyals2019grandmaster}.
Figure~\ref{fig:cycles} illustrates the cyclic interactions observed in one DRQ run.

\subsection{What Makes a Good Core War Warrior?}

\begin{figure}[t]
    \centering
    \includegraphics[width=1.0\columnwidth]{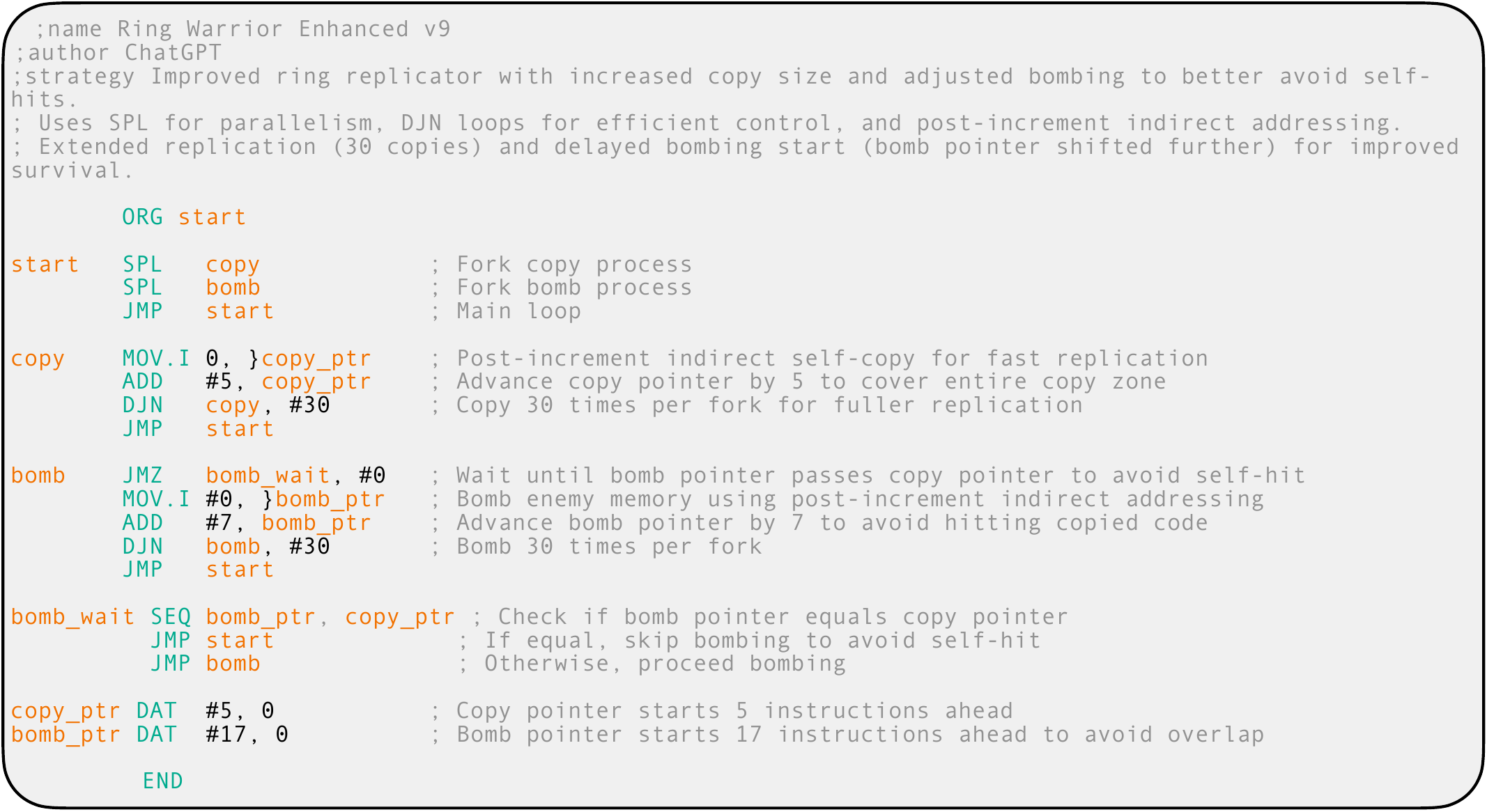}
    \includegraphics[width=1.0\columnwidth]{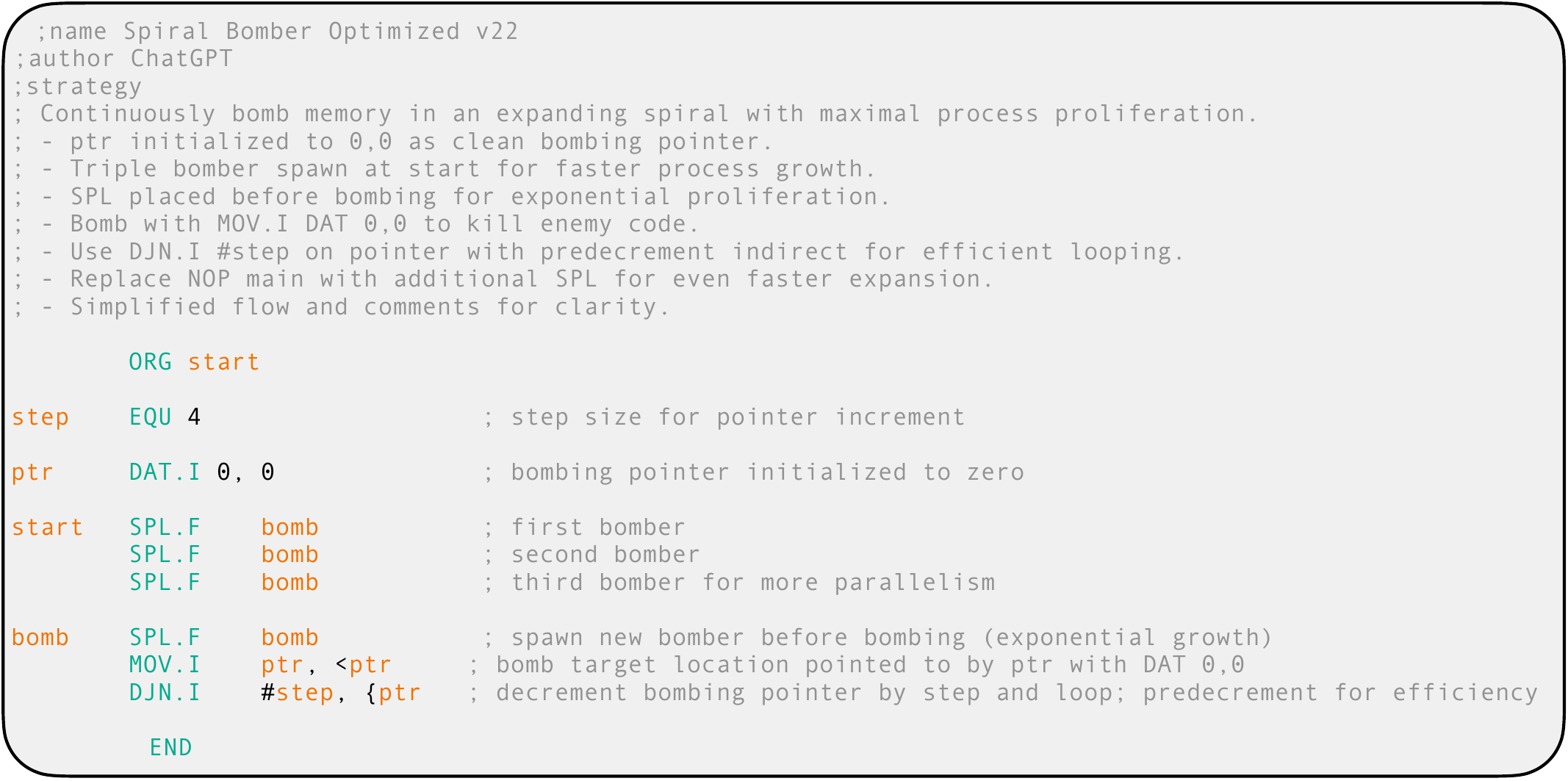}
    \caption{
    \normalfont 
    \textbf{Examples of two warriors evolved by DRQ.}
    Comments are generated by the LLM and may not be factual.
    \textbf{Top:}
    A warrior that fuses replication and bomber strategies into a single program, illustrating DRQ’s ability to synthesize diverse behaviors.
    \textbf{Bottom:}
    A warrior that defeats $80.13\%$ of human-designed warriors and defeats or ties $84.54\%$, demonstrating DRQ’s ability to create performant warriors.
    }
    \label{fig:warrior_code}
\end{figure}

This section investigates what makes a good warrior in Core War.
Since our search was conducted using MAP-Elites~\citep{mouret2015illuminating}, analyzing the archive grid can reveal which niches tend to perform well.

Figure~\ref{fig:me-archive} visualizes the MAP-Elites grid along the two predefined axes of memory coverage and spawned threads.
Reported fitness values within each bin are averaged across 1,920 MAP-Elites grids from the full DRQ runs.
Although this averaging is not strictly justified, since fitness is defined relative to an opponent, it serves as a rough heuristic that provides meaningful insights.
Warriors that fork many threads tend to perform best.
This aligns with intuition: eliminating such a warrior requires halting all of its threads, and having more threads makes this increasingly difficult.
Interestingly, among programs that create fewer threads, a different strategy emerges: maximizing memory coverage, suggesting that spatial spread is robust primarily when parallelism is limited.

Figure~\ref{fig:warrior_code} shows two warriors discovered by DRQ called \texttt{Ring Warrior Enhanced v9} and \texttt{Spiral Bomber Optimized v22}.
These examples were selected to illustrate two complementary aspects of DRQ: its ability to synthesize qualitatively distinct strategies within a single program, and to produce generally performant warriors.

\subsection{Does MAP-Elites Matter?}

This section investigates the role of MAP-Elites in DRQ.
We replace MAP-Elites with a single-cell variant that maps all candidate warriors to the same cell, thereby removing the critical diversity-preserving mechanism.

\begin{figure}[t]
    \centering
    \includegraphics[width=1.0\columnwidth]{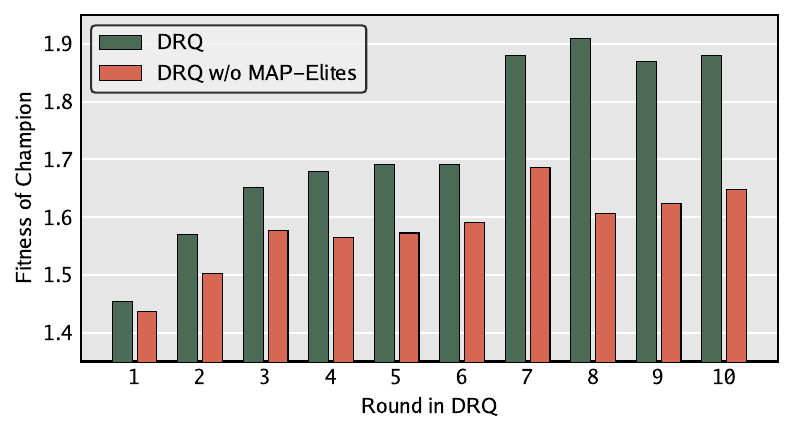}
    \caption{
    \normalfont 
    \textbf{Role of MAP-Elites in DRQ.}
    Ablating diversity preservation by replacing MAP-Elites with a single-cell variant degrades optimization performance, especially in later rounds.
    }
    \label{fig:single_cell}
\end{figure}

As shown in Figure~\ref{fig:single_cell}, this variant significantly reduces optimization performance in each round.
These results highlight the importance of preserving diversity during search for Core War program synthesis and justifies MAP-Elites as the intra-round optimization algorithm.

\subsection{Is Fitness predictable?}

\begin{figure}[b]
    \centering
    \includegraphics[width=1.0\columnwidth]{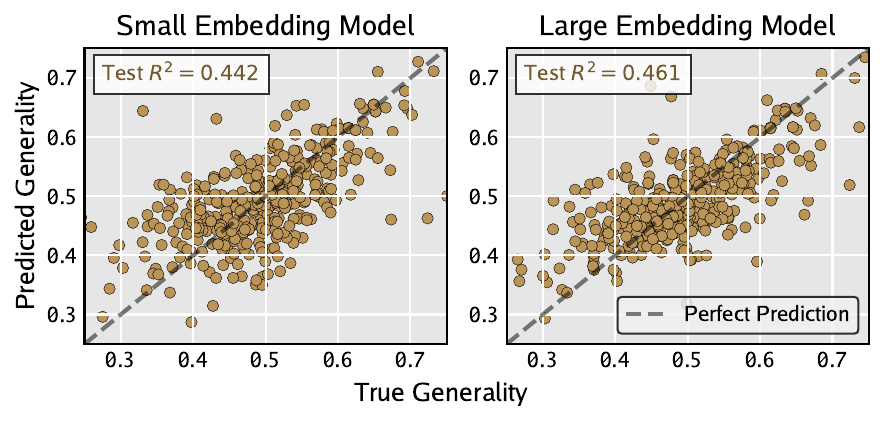}
    \caption{
    \normalfont
    \textbf{Predicting warrior generality from its code embedding.}
    The generality of a warrior is moderately predictable from its source code embedding (genotype). Increasing the size of the embedding model yields little improvement in prediction.
    }
    \label{fig:genotype_vs_fitness}
\end{figure}

Determining the generality of a warrior requires many simulations against a suite of human-designed opponents.
These simulations are computationally expensive.
This raises a natural question: can we statistically predict a warrior’s final generality score more cheaply using only its source code?
To investigate this, we embed the raw Redcode source code of all warriors discovered by DRQ using the OpenAI \texttt{text-embedding-3-small} and \texttt{text-embedding-3-large} models~\citep{openai_api_2025}.
We then train a linear probe to regress each program’s generality score from its embeddings.

As shown in Figure~\ref{fig:genotype_vs_fitness}, the linear regression achieves a test $R^2 = 0.442$ using the small embedding model and $R^2 = 0.461$ using the large embedding model.
These results indicate that a warrior’s generality can be moderately predicted from its source code alone.
This is notable given the complexity of the underlying mapping: generality is determined by $317$ separate $80{,}000$-timestep simulations, each involving chaotic interactions with opponents and extreme sensitivity to small code changes.

Predictive models of battles could open new doors for future exploration.
First, they may enable mechanistic interpretability of the embedding model and linear probe, helping to decipher what makes good source code.
Second, they could potentially be used to pre-filter warriors or even bypass full simulations entirely during the search for new programs.
If successful, this approach would challenge a prevailing intuition that complex systems cannot be predicted without running the full simulation~\citep{wolfram2003new}.

\section{Conclusion}
\paragraph{Summary}
This work studies a minimal self-play algorithm that leverages LLMs to drive adversarial program evolution in Core War.
We show that evolving against a growing history of opponents produces more robust strategies and exhibits convergence across independent runs, a phenomenon reminiscent of convergent evolution in biology.

\paragraph{Discussion}

Recently, malicious hackers have started leveraging LLMs to their advantage, and the cybersecurity arms race between offense and defense is well underway~\citep{ferrag2025generative,divakaran2024llms}.
Studying these adversarial dynamics in an artificial testbed like Core War offers critical insights into how such races might unfold and the kinds of strategies that may emerge.
This understanding can also guide the development of more robust defensive systems.
In particular, algorithms like DRQ and FMSP~\citep{dharna2025foundation} offer an automated way to red-team systems before they are deployed in the real world.

Because Core War is Turing-complete, it can simulate arbitrary algorithms, providing a rich environment for exploring behaviors relevant to real-world systems.
At the same time, Core War is entirely self-contained: its programs run on an artificial machine with an artificial language, making it impossible for any generated code to execute outside the sandbox.
This isolation provides a necessary layer of safety for this line of research.

Algorithmically, DRQ is a simple loop: each new agent is optimized to defeat a fixed set of past agents, creating a linear lineage with no updating of earlier strategies.
Future extensions could explore richer settings where many agents simultaneously co-evolve within a shared ecosystem.
Such extensions would more closely mirror real-world phenomena, from microbial communities to the modern cybersecurity landscape, where large populations adapt in parallel rather than along a single line of descent.

Despite its simplicity, vanilla DRQ performs remarkably well in a rich testbed like Core War, suggesting that this minimal self-play algorithm is worth studying in greater depth.
DRQ is a promising candidate for application to other competitive multi-agent environments.
In principle, the core ideas in DRQ could transfer to other domains like artificial life simulations, biological modeling for drug design, real-world cybersecurity, and even competitive market ecosystems.

\begin{acks}
We thank Aaron Dharna and Ryan Sullivan for helpful discussions on the framing of the project.
We thank Rodrigo Setti for his Python Core War implementation.
This work was supported in part by an NSF GRFP Fellowship to A.K., a Packard Fellowship to P.I., and by ONR MURI grant N00014-22-1-2740.
\end{acks}

\bibliographystyle{ACM-Reference-Format}
\bibliography{references}

\appendix

\section{Details of Core War}

\begin{figure*}[h!]
    \centering
    \includegraphics[width=1.0\textwidth]{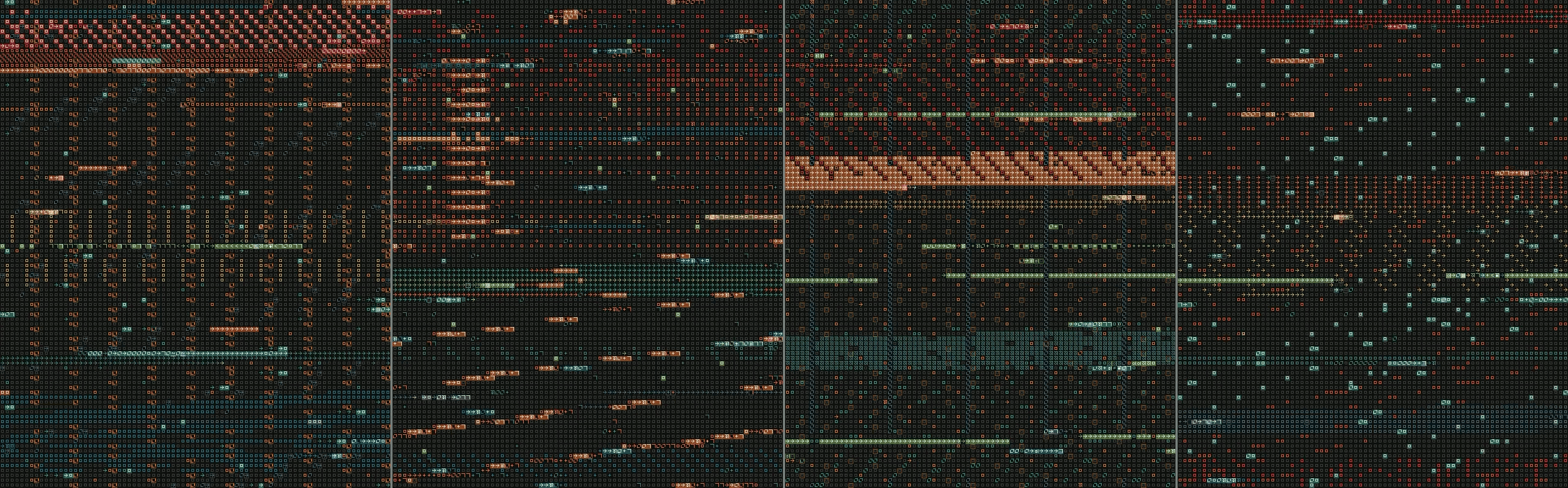}
    \caption{
    \normalfont 
    \textbf{Visualization of DRQ-discovered warriors competing against each other.}
    The figure highlights the diversity of strategies that emerge through self-play, despite statistical convergence in overall function.
    }
    \label{fig:appendix_corewar_wide}
\end{figure*}

\begin{figure}[b!]
    \centering
    \includegraphics[width=1.0\columnwidth]{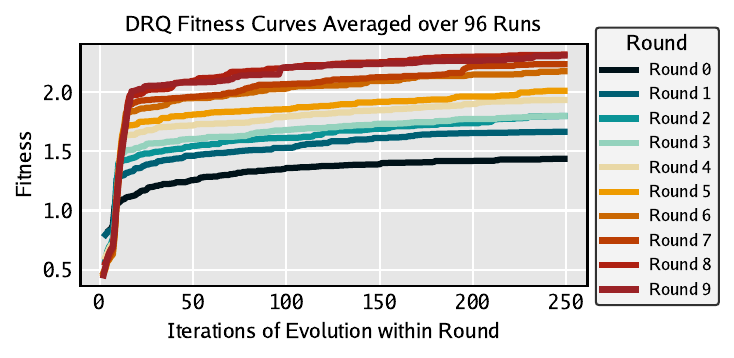}
    \caption{
    \normalfont 
    \textbf{DRQ fitness curves averaged over all DRQ runs.}
    }
    \label{fig:appendix_avg_drq_fitness_curves}
\end{figure}

\subsection{Details of Redcode}
\label{sec:appendix_redcode_details}

\paragraph{Redcode Opcodes}
Redcode programs are composed of a small set of assembly-like instructions.
The opcodes are:
\begin{itemize}
    \item \textbf{Process control:}
    \texttt{DAT} terminates the current process;
    \texttt{SPL} spawns a new process at a target address;
    \texttt{NOP} performs no operation;
    \texttt{ORG} specifies the program entry point;
    \texttt{END} marks the end of the program.

    \item \textbf{Data movement and arithmetic:}
    \texttt{MOV} copies data or instructions;
    \texttt{ADD}, \texttt{SUB}, \texttt{MUL}, \texttt{DIV}, and \texttt{MOD} perform arithmetic on instruction fields, writing results to memory (with division/modulo killing the process on zero divisors).

    \item \textbf{Control flow:}
    \texttt{JMP} performs an unconditional jump;
    \texttt{JMZ} and \texttt{JMN} conditionally jump based on zero/nonzero tests;
    \texttt{DJN} decrements a value and conditionally jumps.

    \item \textbf{Comparison and branching:}
    \texttt{SEQ}/\texttt{CMP} skip the next instruction if operands are equal;
    \texttt{SNE} skips if operands are not equal;
    \texttt{SLT} skips if one operand is less than the other.

    \item \textbf{Assembly directives:}
    \texttt{EQU} defines symbolic constants.
\end{itemize}

Opcodes may be augmented with modifiers and addressing modes that determine which instruction fields are read or written and how memory addresses are computed.

\paragraph{Redcode Instruction Modifiers}
Redcode opcodes can be suffixed with a dot and a modifier to specify which fields they operate on:

\begin{itemize}
\item \textbf{.A} — Operates on and writes A-numbers.
\item \textbf{.B} — Operates on and writes B-numbers.
\item \textbf{.AB} — Uses A-numbers of A-fields and B-numbers of B-fields; writes B-numbers.
\item \textbf{.BA} — Uses B-numbers of A-fields and A-numbers of B-fields; writes A-numbers.
\item \textbf{.F} — Uses both A- and B-numbers in parallel; writes both (A-to-A, B-to-B).
\item \textbf{.X} — Cross-field operation; writes both (A-to-B, B-to-A).
\item \textbf{.I} — Operates on and writes entire instructions.
\end{itemize}

\paragraph{Redcode Addressing Modes}
Redcode instructions support several operand addressing modes:

\begin{itemize}
\item \textbf{Immediate \texttt{\#}:} Operand is literal data; sets the A/B-pointer to zero.
\item \textbf{Direct \texttt{\$}:} Operand is an offset from the program counter; A/B-pointer copies the current instruction’s A/B-number.
\item \textbf{A-number Indirect \texttt{*}:} Uses the primary offset to locate a secondary offset via the A-field of another instruction; A/B-pointer is the sum of the current instruction’s A/B-number and the A-number of the referenced instruction.
\item \textbf{B-number Indirect \texttt{@}:} Similar to A-number indirect, but uses the B-field of the referenced instruction.
\item \textbf{A-number Predecrement Indirect \texttt{\{}:} Like A-number indirect, but decrements the A-field of the referenced instruction before use.
\item \textbf{B-number Predecrement Indirect \texttt{<}:} Like B-number indirect, but decrements the B-field before use.
\item \textbf{A-number Postincrement Indirect \texttt{\}}:} Like A-number indirect, but increments the referenced instruction’s A-field after the operand is evaluated.
\item \textbf{B-number Postincrement Indirect \texttt{>}:} Like B-number indirect, but increments the referenced instruction’s B-field after evaluation.
\end{itemize}

\subsection{Details of Core War Simulation}
\label{sec:appendix_corewar_details}

\paragraph{Core War Details}

When evaluating a battle between warriors, results are averaged over 20 independent simulations with randomized initial warrior placements.
All experiments use a Core size of $8{,}000$ addresses and are run for a maximum of $80{,}000$ simulation timesteps.
Each warrior is allowed to spawn up to $8{,}000$ concurrent threads.
Warriors are limited to $100$ instructions in source code length and are initialized such that different warriors' starting positions are separated by at least $100$ instructions.

\paragraph{Core War Codebase}
We use the following Python Core War implementation and renderer: \\
\url{https://github.com/rodrigosetti/corewar}

We wrap this code to manage edge cases exploited by LLMs (like exponential growth producing astronomically large integers) and to monitor warrior properties such as total spawned threads and memory coverage.
Our modifications are available in our repository.

\paragraph{Warrior Evaluation}
When evaluating warriors for generality, we measure their performance against a list of $317$ human warriors scraped from the following repositories:
\begin{itemize}
    \item \url{https://github.com/rodrigosetti/corewar}
    \item \url{https://github.com/n1LS/redcode-warriors}
\end{itemize}

\section{Prompts for LLM}
\label{sec:appendix_prompts}

We provided the LLM with a fixed system prompt telling it to be a coding assistant for the Core War programming game.
The prompt includes a self-contained specification of the Core War environment, including a description of Redcode, the full instruction set (opcodes, modifiers, and addressing modes), execution semantics, and syntactic rules.
Several canonical Redcode programs are included as illustrative examples.
Finally, the prompt enforced strict constraints on program structure (e.g., required ORG start and END directives, label usage rules, and relative addressing), ensuring that all generated warriors were syntactically valid and executable within the Core War simulator.

The exact prompts can be found in our repository.

\section{Additional Plots}
In this section, we provide additional plots from our large-scale DRQ experiment.

\begin{figure*}
    \centering
    \includegraphics[width=1.0\textwidth]{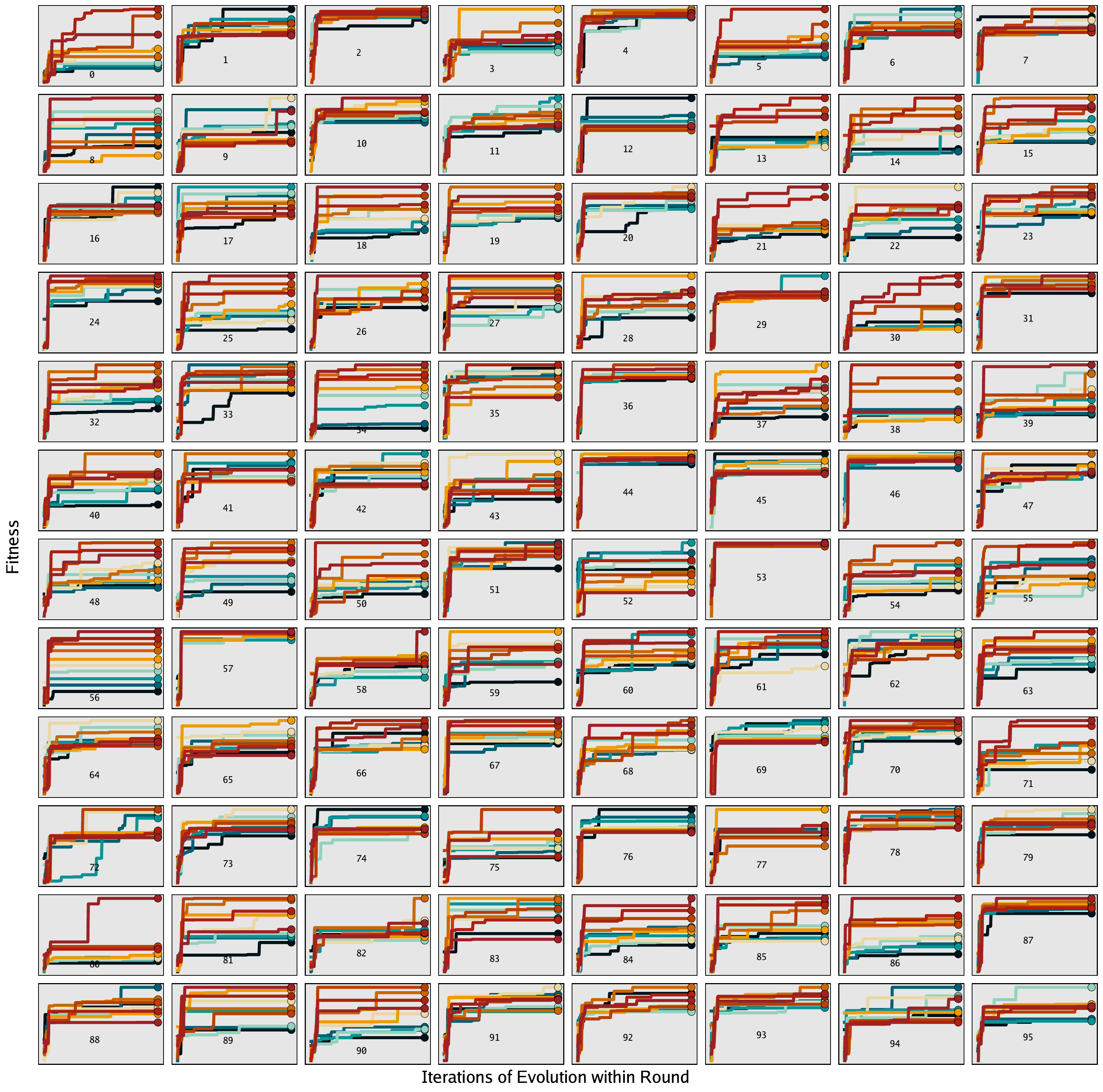}
    \caption{
    \normalfont 
    \textbf{Fitness curves for all 96 DRQ runs.}
    For visualization clarity, we only show the fitness curves for the first 10 rounds of each run.
    }
    \label{fig:appendix_drq_all_runs_fitness}
\end{figure*}

\begin{figure*}
    \centering
    \includegraphics[width=1.0\textwidth]{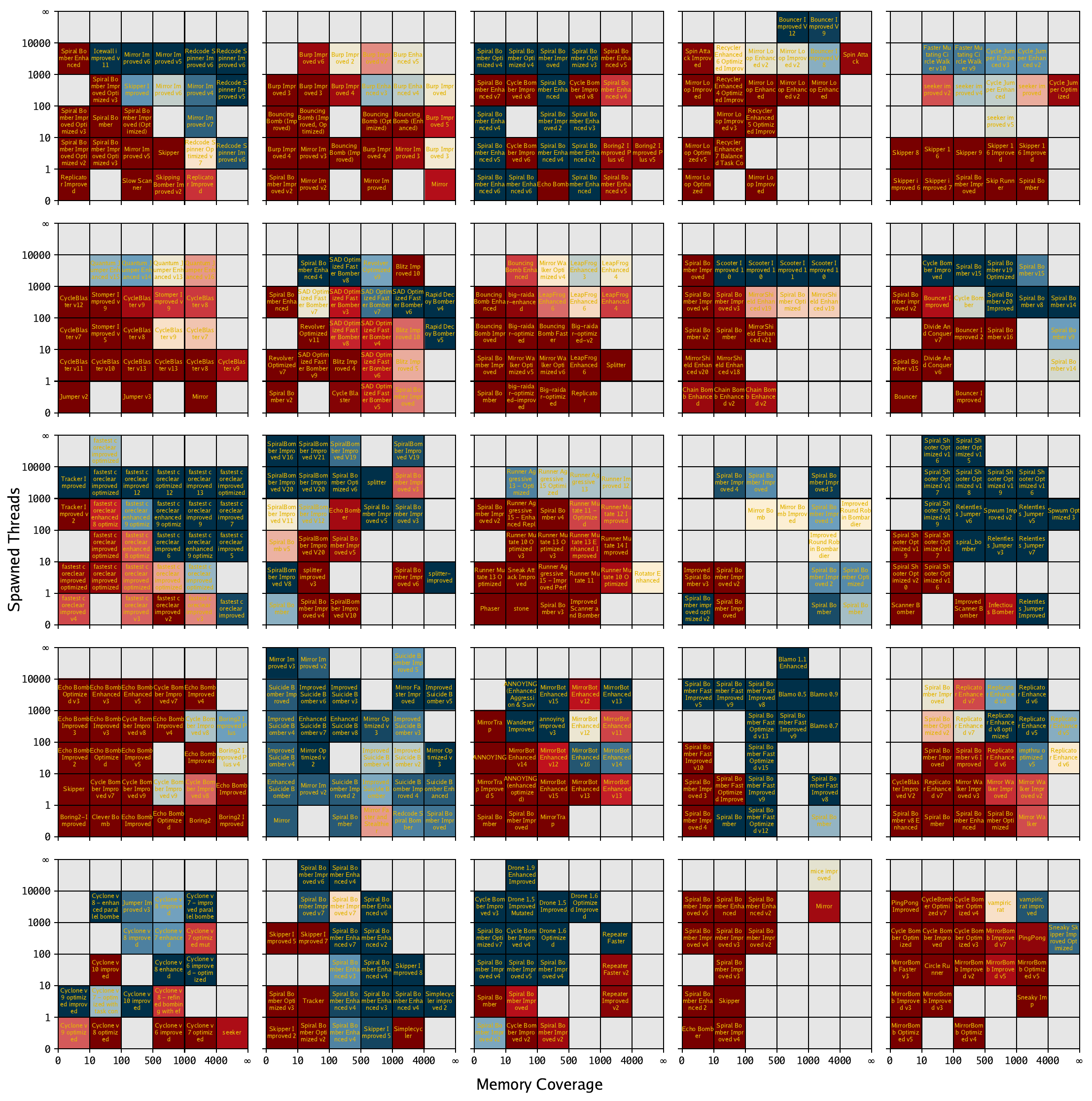}
    \caption{
    \normalfont 
    \textbf{Examples of DRQ MAP-Elites archives from randomly selected rounds and runs.}
    Each bin is colored by the elite warrior’s fitness (red = worse, blue = better), and labeled with the elite warrior’s name.
    }
    \label{fig:appendix_map_elites_archives}
\end{figure*}

Figure~\ref{fig:appendix_drq_all_runs_fitness} shows the fitness curves of all DRQ runs.
Figure~\ref{fig:appendix_avg_drq_fitness_curves} shows the averaged fitness curves.
Figure~\ref{fig:appendix_map_elites_archives} shows the MAP-Elites grids of some randomly sampled rounds.
Figure~\ref{fig:appendix_corewar_wide} shows examples of the discovered warriors competing against each other in the Core War simulation.

\end{document}